\pdfoutput=1
\documentclass[11pt]{article}

\usepackage[preprint]{acl}
\usepackage{times}
\usepackage{latexsym}

\usepackage[T1]{fontenc}
\usepackage[utf8]{inputenc}

\usepackage{microtype}

\usepackage{inconsolata}

\usepackage{graphicx}

\usepackage{booktabs}
\usepackage{tabularx}
\usepackage{float}
\usepackage{multirow}

\usepackage{enumitem}
\usepackage{tcolorbox}
\usepackage{tikz}
\usepackage{pgfplots}

\definecolor{uzhapple}{RGB}{164,210,51}
\definecolor{uzhblue}{RGB}{0,40,165}
\definecolor{graycolor}{rgb}{0.6,0.6,0.6}

\title{Expanding the WMT24++ Benchmark with Rumantsch Grischun, \\Sursilvan, Sutsilvan, Surmiran, Puter, and Vallader}

\author{
  \textbf{Jannis Vamvas\textsuperscript{1}}\quad
  \textbf{Ignacio Pérez Prat\textsuperscript{2}}\quad
  \textbf{Not Battesta Soliva\textsuperscript{1}}\quad \\
    \textbf{Sandra Baltermia-Guetg\textsuperscript{2}}\quad
    \textbf{Andrina Beeli\textsuperscript{2}}\quad
    \textbf{Simona Beeli\textsuperscript{2}}\quad
    \textbf{Madlaina Capeder\textsuperscript{2}}\quad  \\
    \textbf{Laura Decurtins\textsuperscript{2}}\quad
    \textbf{Gian Peder Gregori\textsuperscript{2}}\quad
    \textbf{Flavia Hobi\textsuperscript{2}}\quad
    \textbf{Gabriela Holderegger\textsuperscript{2}}\quad \\
    \textbf{Arina Lazzarini\textsuperscript{2}}\quad
    \textbf{Viviana Lazzarini\textsuperscript{2}}\quad
    \textbf{Walter Rosselli\textsuperscript{2}}\quad
    \textbf{Bettina Vital\textsuperscript{2}}\quad \\
  \textbf{Anna Rutkiewicz\textsuperscript{1}}\quad
  \textbf{Rico Sennrich\textsuperscript{1}}
\\[0.3em]
  \textsuperscript{1}University of Zurich \quad
  \textsuperscript{2}Lia Rumantscha
\\
  \small{
    Correspondence: \href{mailto:vamvas@cl.uzh.ch}{vamvas@cl.uzh.ch}, \href{mailto:ignacio.perez.prat@rumantsch.ch}{ignacio.perez.prat@rumantsch.ch}
  }
}

\begin{document}

%\AddToHook{shipout/foreground}{
%    \begin{tikzpicture}[overlay, remember picture]
%        \node[opacity=0.12, text=gray, rotate=45, scale=8] at (current page.center) {DRAFT};
%    \end{tikzpicture}
%}

\maketitle
\begin{abstract}
The Romansh language, spoken in Switzerland, has limited resources for machine translation evaluation.
In this paper, we present a benchmark for six varieties of Romansh: Rumantsch Grischun, a supra-regional variety, and five regional varieties: Sursilvan, Sutsilvan, Surmiran, Puter, and Vallader.
Our reference translations were created by human translators based on the WMT24++ benchmark, which ensures parallelism with more than 55 other languages.
An automatic evaluation of existing MT systems and LLMs shows that translation out of Romansh into German is handled relatively well for all the varieties, but translation into Romansh is still challenging.

%\vspace{0.5em}
%\begin{center}
%  \setlength{\fboxsep}{6pt}
%  \definecolor{customblue}{HTML}{E3EAF7}
%  \colorbox{customblue}{
%    \begin{tabular}{ll}
%      \textbf{ISO 639-1:} & \texttt{rm} \\
%      \textbf{ISO 639-2/3:} & \texttt{roh} \\
%    \end{tabular}
%  }
%\end{center}
%\vspace{-.5em}
\end{abstract}

\section{Introduction}
The automatic evaluation of machine translation~(MT) has been widened in recent years to cover more languages and language varieties.
While massively multilingual benchmarks such as FLORES~\cite{goyal-etal-2022-flores, Costa-jussà2024} or NTREX~\cite{federmann-etal-2022-ntrex} include reference translations in hundreds of languages, no dedicated reference translations for the Romansh language have been available so far.
In this paper, we close this gap by extending the recent WMT24++ benchmark~\cite{kocmi-etal-2024-findings, deutsch2025wmt24expandinglanguagecoverage} with reference translations for six varieties of Romansh, using German as the source language.

\begin{figure}[ht]
    \centering
    \includegraphics[width=\columnwidth,trim={0 0 85 0},clip]{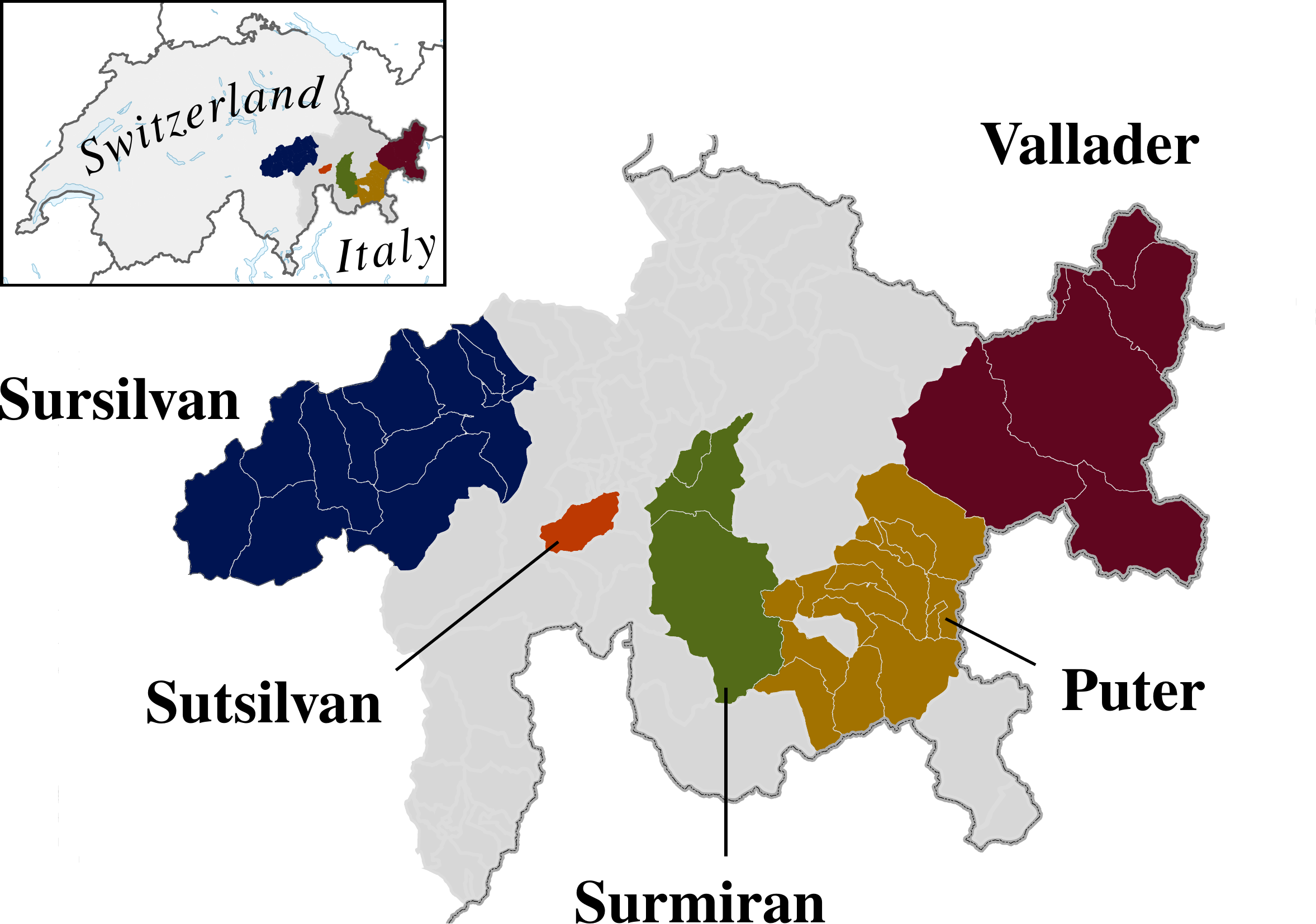}
    \caption{Distribution of Romansh \textit{idioms} (regional varieties) within south-eastern Switzerland.
    The map shows municipalities where an idiom is officially used in public administration.
    We extend the WMT24++ benchmark with sets of reference translations for these five idioms, as well as Rumantsch Grischun, a supra-regional variety of Romansh.
    }
    \label{fig:idiom-map}
\end{figure}

There are several reasons why Romansh, which is a language from the Romance family spoken in Switzerland (\textbf{ISO 639-1:~\texttt{rm}; ISO 639-2/3:~\texttt{roh}}), has had limited resources for MT evaluation.
First, Romansh is considered a minority language, with 40,000–60,000 speakers~\cite{gross_romansh_2004,grunert2024raetoromanisch}.
Secondly, multiple varieties of Romansh need to be considered for a comprehensive evaluation.
\textit{Rumantsch Grischun} is a supra-regional variety of the language, often used in official contexts.
However, the five regional varieties of Romansh~(Figure~\ref{fig:idiom-map}), usually referred to as \textit{idioms}, are more widely spoken in everyday life, with limited mutual intelligibility~\cite{gross_romansh_2004}.

Prior work on MT for Romansh~\cite{muller-etal-2020-domain,niklaus2025swiltra} has leveraged multilingual government press releases~\cite{scherrer-cartoni-2012-trilingual}, blog posts or federal laws, all of which cover only the Rumantsch Grischun variety.
Our benchmark based on WMT24++ enables a more systematic evaluation setup that includes the five idioms and a broader range of domains, such as social media and transcripts of YouTube videos.

% Provide a short excerpt of the data available in the dataset to demonstrate its format and content. This should take up no more than half a page. If possible, provide a translation in English.
\begin{table*}[ht]
\centering
\begin{tabularx}{\textwidth}{@{}lX@{}}
& \textbf{Data Sample} \\
\toprule
\multirow{1}{*}{\shortstack[l]{\textbf{English}~\cite{kocmi-etal-2024-findings}}}
    & \textit{it seems like even iMessage over WiFi isn't working, which doesn't quite make sense to me} \\
\midrule
\multirow{1}{*}{\shortstack[l]{\textbf{German}~\cite{deutsch2025wmt24expandinglanguagecoverage}}}
    & \textit{Anscheinend funktioniert nicht mal iMessage über WiFi, was mir nicht ganz einleuchtet} \\
\toprule
\multirow{2}{*}{\shortstack[l]{\textbf{Rumantsch Grischun}\\[0.3em]Code: \texttt{\textbf{roh\_Latn\_ruma1247}}}}
    & \textit{Para che gnanc iMessage funcziunia via WiFi, tge ch'è per mai betg dal tut evident.} \\
\midrule
\multirow{2}{*}{\shortstack[l]{\textbf{Sursilvan}\\[0.3em]Code: \texttt{\textbf{roh\_Latn\_surs1244}}}}
    & \textit{Sco ei para funcziunescha gnanc iMessage sur WiFi, quei ch'jeu sai buca propi capir.} \\
\midrule
\multirow{2}{*}{\shortstack[l]{\textbf{Surmiran}\\[0.3em]Code: \texttt{\textbf{roh\_Latn\_surm1243}}}}
    & \textit{Scu para funcziunescha mianc iMessage sur WiFi, chegl tg'ia sa betg propi tgapeir} \\
\midrule
\multirow{2}{*}{\shortstack[l]{\textbf{Sutsilvan}\\[0.3em]Code: \texttt{\textbf{roh\_Latn\_suts1235}}}}
    & \textit{Para funcziunescha gnànc iMessage sur igl WiFi, tge ca fa betga propi sen tanor me} \\
\midrule
\multirow{2}{*}{\shortstack[l]{\textbf{Puter}\\[0.3em]Code: \texttt{\textbf{roh\_Latn\_uppe1396}}}}
    & \textit{Pera cha nu funcziuna niauncha iMessage sur WiFi, che ch'eau nun incleg dal tuot} \\
\midrule
\multirow{2}{*}{\shortstack[l]{\textbf{Vallader}\\[0.3em]Code: \texttt{\textbf{roh\_Latn\_lowe1386}}}}
    & \textit{Apparaintamaing nu funcziuna gnanca üna jada iMessage sur WiFi, quai chi nu'm voul propcha ir per testa} \\
\bottomrule
\end{tabularx}
\caption{Samples of the Romansh varieties that we contribute to the benchmark, plus the English and German segments from prior work. The samples are from the \textit{Social} domain. The language code assigned by the Open Language Data Initiative (OLDI) has three components: the ISO~639-3 language code (\texttt{roh}), the ISO~15924 script code (\texttt{Latn}), and the Glottocode assigned to the variety by Glottolog~\cite{glottolog2025}.}
\label{tab:literary-sample}
\end{table*}

We release our benchmark under the Apache~2.0 license.\footnote{\url{https://hf.co/datasets/ZurichNLP/wmt24pp-rm}}
In addition, we use the benchmark to perform a systematic evaluation of MT systems and LLMs on German--Romansh and Romansh--German translation.
Results based on automatic evaluation metrics indicate that translation into German achieves reasonable quality for all Romansh varieties, while translation into Romansh remains challenging, particularly for the less-resourced idioms.
Code for reproducing our experiments is available.\footnote{\url{https://github.com/ZurichNLP/romansh_mt_eval}}

\section{Language Overview}

\subsection{Romansh}
Romansh is part of the Romance branch of the Indo-European language family. It is a minority language in the Swiss canton of Graubünden and is treated as one of the country's four national languages \citep{grunert_multilingualism_2018}. Its status is considered endangered \citep{moseley_atlas_2010}.
% (Moseley/Nicolas 2010: 24). 
Romansh covers an extremely diverse dialect continuum spanning the canton, where roughly 15\% of inhabitants speak it as their main language \citep{gross_romansh_2004}.
% (Gross 2004: 26).
Unlike other dialect continuums, Romansh is not ``roofed'' by a single standard language \citep{goebl2003externe}.
% (Goebl 2003: 751-753). 
Instead, there are five different written traditions dividing the Romansh-speaking area into regions with their own written standards---known as \textit{idioms}---that differ heavily from each other in all areas of language structure \citep{liver_ratoromanisch_2010,haiman_rhaeto-romance_1992,schmid_zur_1976}.
% (consider Liver 2010, Haiman/Benincà 1992, Schmid 1976). 
% These are referred to as \textit{idioms} and they are written forms of the dialects they cover.
In the 1980s, linguist Heinrich Schmid developed Rumantsch Grischun, a supra-regional, constructed standard, as a \textit{Dachsprache}
% (Muljačić 1989) 
for Romansh \citep{muljacic_uber_2012}.
% We will discuss first the \textit{idioms} from west to east and then Rumantsch Grischun. 

% Census data is important to gain an overview of Romansh's status, but it is far from reliable  \citep{coray_struggle_2020}. The data presented here from 2000 identifies the number of speakers in each idiom's territory, but this does not include the roughly 26,500 Romansh speakers living outside the traditional territory. The speaker estimates are therefore conservative. The speaker numbers are historically stable \citep{obrist_discourse_2022}.
% (Obrist 2022: 615).
% and Romansh speakers also know German. 

\subsection{Sursilvan}
Sursilvan is used in the west of Graubünden, in an area mainly covered by the Surselva valley. In many municipalities towards the east, it is still the predominant first language, while German is increasingly dominant moving west \citep{gross_romansh_2004}.
% (Gross 2004: 33). 
Sursilvan is the idiom with the largest population. The written form mainly represents the dialects spoken between Disentis and Ilanz, though the entire area it covers is a continuum exhibiting mutual intelligibility. There were at least 18,000 Sursilvan speakers in the year 2000 \citep{gross_romansh_2004}.
% (Gross 2004: 31)

\subsection{Sutsilvan}
Sutsilvan is spoken in the valley of the Hinterrhein river, though its territory is no longer contiguous. It is the variety with the lowest number of speakers and the highest level of endangerment \citep{liver_romanche_2014}.
% (Liver 2014: 414). 
Large parts of Sutsilvan's traditional speaker territory became German speaking several centuries ago, and until recently, there was no established written form for Sutsilvan.
%and little interest in maintaining the variety.
A concentrated effort to change this was initiated by Giuseppe Gangale in the 1940s, establishing modern Sutsilvan orthography. His approach, however, sparked debate,
% (Coray 2008: 117-123) 
and Sutsilvan remains an idiom with hardly any majority Romansh territory \citep{coray_von_2008}. There were at least 1,000 Sutsilvan speakers in the year 2000 \citep{gross_romansh_2004}.
% (Gross 2004: 31)

\subsection{Surmiran}
Surmiran is spoken in central Graubünden, namely in the regions of Alvra/Sotses and Surses, the latter being an area where Romansh is still largely present in everyday life \citep{liver_romanche_2014}.
% (Liver 2014: 414). 
Together with Sutsilvan, Surmiran constitutes a bridge between the starkly different dialects of the Surselva region in western Graubünden and the Engadine valley in the east. For this reason, it has previously been suggested as a lingua franca for supra-regional communication \citep{coray_von_2008}.
% (Coray 2008:127). 
Surmiran itself exhibits some peculiarities, however, shared by neither of the two other major Romansh-speaking areas. There were at least 3,000 Surmiran speakers in the year 2000 \citep{gross_romansh_2004}.
% (Gross 2004: 31)

\subsection{Puter}
Puter and Vallader are used as written standards in the Engadine valley, with Puter being used south of Zuoz. The Engadine valley can itself be seen as a continuum of varieties more diverse than the Surselva \citep{schmid_zur_1976}.
% (Schmid 1976: 17). 
Puter reflects characteristics of the dialects in the upper Engadine valley, with more Italian influence than Vallader. Written Puter dates back to 1552 \citep{obrist_discourse_2022},
% (Obrist 2022: 618), 
and is thus the variety with the longest-standing written tradition. Puter is under substantial pressure from German due to growing tourism since the last century \citep{liver_romanche_2014}.
% (Liver 2014: 415). 
Municipalities with a Romansh majority have become scarce. There were at least 5,500 Puter speakers in the year 2000 \citep{gross_romansh_2004}.
% (Gross 2004: 31)

\subsection{Vallader}
Vallader is used in the Lower Engadine valley, north of Zernez, as well as in the Val Müstair.
% even though the latter has a rather distinct spoken variety. 
Vallader, unlike Puter, remains a majority language in most of its territory \citep{liver_romanche_2014}.
% (Liver 2014: 415). 
Together with Sursilvan, the Vallader territory represents a stronghold of Romansh. There were at least 6,500 Vallader speakers in the year 2000 \citep{gross_romansh_2004}.
% (Gross 2004: 31)

\subsection{Rumantsch Grischun}
The Rumantsch Grischun variety has a special role in that it is not an idiom, but a written standard devised as a constructed language. It does not reflect any Romansh speaker's natural speech, but was constructed to be a globally intelligible and neutral written form that could be used to represent Romansh as a language. It was developed by comparing structural and lexical characteristics of the different idioms and determining the most mutually intelligible forms
% and settle on these as a basis for a new language, a smallest common denominator of sorts 
\citep{schmid_richtlinien_1982}.
% (Schmid 1982). 

% This attempt to provide the hitherto missing \textit{Dachsprache} for the Romansh idioms was successful in some ways. 
Rumantsch Grischun is used for official publications from the canton or the federal government, as well as other institutions addressing the entire Romansh population. More extended promotion of Rumantsch Grischun (including replacing the idioms as the language of literacy at schools)
% by the Lia Rumantscha and the canton 
met heavy resistance 
% (see Coray 2008)
and caused long-lasting debate \citep{coray_von_2008}.
% Teaching Rumantsch Grischun instead of the idioms at school has been abandoned in the few municipalities that tentatively adopted it.
Most speakers of Romansh only actively learn their own idiom. Though they may occasionally come into contact with Rumantsch Grischun texts, their knowledge of it is only passive at most.

\section{Data Collection}

\subsection{Choice of Benchmark}
We chose to extend the WMT24++ benchmark~\cite{kocmi-etal-2024-findings,deutsch2025wmt24expandinglanguagecoverage} based on the following considerations:
\begin{itemize}[topsep=0.4em, itemsep=0em]
  \item WMT24++ currently covers 55 languages, including the other Swiss national languages (German, French and Italian).
  % as well as the following additional Romance languages: Spanish (Mexico), Catalan, French (Canada), Portuguese (Brazil), Portuguese (Portugal), and Romanian.
  \item It is a recent benchmark that is unlikely to suffer from data contamination in LLMs.
  \item Segments are provided in context, allowing for document-level evaluation.
\end{itemize}

\begin{figure*}[t]
\centering
\begin{minipage}[t]{\columnwidth}
\centering

\begin{tabularx}{\columnwidth}{@{}X|*{6}{>{\raggedleft\arraybackslash}p{0.65cm}}}
\multirow[b]{2}{*}{$\downarrow$~\textbf{pred}}
& \multicolumn{6}{l}{$\textbf{gold}$~$\rightarrow$} \\
& RG & Surs.\ & Suts.\ & Surm.\ & Puter & Vall. \\
\midrule
RG    & \cellcolor{uzhapple!69.0}764   & \cellcolor{uzhapple!7.5}83   & \cellcolor{uzhapple!1.5}23   & \cellcolor{uzhapple!3.75}43   & \cellcolor{uzhapple!2.25}29   & \cellcolor{uzhapple!3.0}35   \\
Surs.  & \cellcolor{uzhapple!8.25}98 & \cellcolor{uzhapple!73.5}810 & \cellcolor{uzhapple!6.0}67 & \cellcolor{uzhapple!3.75}46 & \cellcolor{uzhapple!4.5}55 & \cellcolor{uzhapple!4.5}54 \\
Suts.  & \cellcolor{uzhapple!0.0}8 & \cellcolor{uzhapple!0.75}12 & \cellcolor{uzhapple!73.5}809 & \cellcolor{uzhapple!1.5}20 & \cellcolor{uzhapple!0.0}6 & \cellcolor{uzhapple!0.75}10 \\
Surm. & \cellcolor{uzhapple!1.5}19 & \cellcolor{uzhapple!0.75}9 & \cellcolor{uzhapple!2.25}32 & \cellcolor{uzhapple!73.5}811 & \cellcolor{uzhapple!0.0}8 & \cellcolor{uzhapple!0.0}6 \\
Puter    & \cellcolor{uzhapple!0.75}10   & \cellcolor{uzhapple!0.75}12   & \cellcolor{uzhapple!0.75}12   & \cellcolor{uzhapple!1.5}17   & \cellcolor{uzhapple!58.5}648   & \cellcolor{uzhapple!2.25}30   \\
Vall.  & \cellcolor{uzhapple!5.25}61 & \cellcolor{uzhapple!3.0}34 & \cellcolor{uzhapple!1.5}17 & \cellcolor{uzhapple!1.5}23 & \cellcolor{uzhapple!18.75}214 & \cellcolor{uzhapple!75.0}825 \\
\bottomrule
\end{tabularx}

\captionof{table}{Confusion matrix of a Romansh language variety classifier when applied to the reference translations.}
\label{tab:language-classification}
\end{minipage}%
\hfill
\begin{minipage}[t]{\columnwidth}
\centering

\begin{tabularx}{\columnwidth}{@{}X|*{6}{>{\raggedleft\arraybackslash}p{0.65cm}}}
\multirow[b]{2}{*}{$\downarrow$~\textbf{sys}}
& \multicolumn{6}{l}{$\textbf{ref}$~$\rightarrow$} \\
& RG & Surs.\ & Suts.\ & Surm.\ & Puter & Vall. \\
\midrule
RG    & \cellcolor{uzhapple!75.0}   & \cellcolor{uzhapple!26.25}60.0   & \cellcolor{uzhapple!10.5}47.5   & \cellcolor{uzhapple!18.75}53.8   & \cellcolor{uzhapple!10.5}47.2   & \cellcolor{uzhapple!12.75}49.4   \\
Surs.  & \cellcolor{uzhapple!27.0}60.7 & \cellcolor{uzhapple!75.0} & \cellcolor{uzhapple!19.5}54.5 & \cellcolor{uzhapple!13.5}49.5 & \cellcolor{uzhapple!5.25}43.1 & \cellcolor{uzhapple!5.25}43.0 \\
Suts.  & \cellcolor{uzhapple!12.0}48.3 & \cellcolor{uzhapple!19.5}54.7 & \cellcolor{uzhapple!75.0} & \cellcolor{uzhapple!15.0}50.9 & \cellcolor{uzhapple!0.75}39.5 & \cellcolor{uzhapple!0.75}39.1 \\
Surm. & \cellcolor{uzhapple!19.5}54.8 & \cellcolor{uzhapple!13.5}49.8 & \cellcolor{uzhapple!15.0}51.0 & \cellcolor{uzhapple!75.0} & \cellcolor{uzhapple!5.25}43.1 & \cellcolor{uzhapple!5.25}43.3 \\
Puter    & \cellcolor{uzhapple!10.5}47.4   & \cellcolor{uzhapple!5.25}42.7   & \cellcolor{uzhapple!0.0}39.0   & \cellcolor{uzhapple!4.5}42.4   & \cellcolor{uzhapple!75.0}   & \cellcolor{uzhapple!24.0}58.7   \\
Vall.  & \cellcolor{uzhapple!12.75}49.3 & \cellcolor{uzhapple!4.5}42.4 & \cellcolor{uzhapple!0.0}38.4 & \cellcolor{uzhapple!4.5}42.4 & \cellcolor{uzhapple!24.0}58.4 & \cellcolor{uzhapple!75.0} \\
\bottomrule
\end{tabularx}

\captionof{table}{Pairwise ChrF scores between the reference translations for the different varieties.}
\label{tab:pairwise-chrf}
\end{minipage}
\end{figure*}

\subsection{Creation of Reference Translations}

% Offer an in-depth description of the data acquisition process. For example, when submitting a translated dataset, provide details such as the source language, number of translators, their expertise (native speakers, proficiency, professional experience), and whether any portion was independently reviewed by third parties.

The data acquisition process was structured into three steps to ensure high quality, consistency, and adherence to idiom-specific conventions.
\begin{enumerate}[topsep=0.4em, itemsep=0em]
  \item \textbf{Translation}: We hired language professionals who are native speakers of both German and the respective Romansh idiom.
  \item \textbf{Review}: Two expert linguists of Lia Rumantscha reviewed a sample of translations for a representative selection of varieties, and formulated feedback that was communicated to all translators.
  \item \textbf{Revision}: The translators incorporated the feedback into the reference translations.
\end{enumerate}

\noindent{}We provided the translators with a guidelines document, inspired by the WMT24 translator brief~\cite{kocmi-etal-2024-findings}.
The key points of the guidelines, which we provide in Appendix~\ref{sec:guidelines}, are:
\begin{itemize}[topsep=0.4em, itemsep=0em]
  \item The German text is the main source for the translation into Romansh, while the English text can be used as an additional reference in case of ambiguity.
  \item No AI tools should be used for the translation.
\end{itemize}

\noindent{}The translators and reviewers had access to the complete context of each segment, including a link to the original website from which the segment was extracted (e.g., for segments from the Speech domain, the original YouTube video).

\subsection{Challenges in the Data Acquisition} A challenge we encountered in the translation process was that the degree of standardization can vary across text domains.
The Romansh idioms are well-standardized, which is reflected in the formal domains \textit{News} and (partially) \textit{Literary}.
However, in the \textit{Social} and \textit{Speech} domains, there is more room for individual variation based on the translator's style or regional background.
Therefore, while we consider the reference translations suitable for their intended use of evaluating idiom-aware MT, the dataset does not aim to represent the full spectrum of variation present in the Romansh idioms.

\section{Validation Experiments}

We perform two automatic validation experiments to confirm that the reference translations are suitable for variety-specific evaluation:

\paragraph{Language Classification}
We use a fastText classifier~\cite{joulin-etal-2017-bag} trained on a corpus of Romansh newspaper articles that were manually labeled with their variety.
Table~\ref{tab:language-classification} shows that when applied to our reference translations, the classifier predicts the expected variety for the majority of segments, indicating that the reference translations exhibit variety-specific features.

\paragraph{Cross-Variety Scores}
We calculate pairwise ChrF scores~\cite{popovic-2015-chrf} between the sets of references, which are reported in Table~\ref{tab:pairwise-chrf}.
The maximum ChrF score across varieties is 60.7~(for Sursilvan–Rumantsch Grischun), which confirms that the sets of reference translations are distinct from each other even for related varieties, allowing for variety-specific evaluation.
At the same time, the cross-variety scores are high enough to rule out serious data quality issues, such as a systematic misalignment of segments.

\section{Evaluation of MT Systems and LLMs on Translation from and into Romansh}

\begin{table*}[ht]
\centering

\begin{tabularx}{\textwidth}{@{}Xrrrrrr@{}}
\toprule
\textbf{System} & \textbf{Rumantsch Grischun} & \textbf{Sursilvan} & \textbf{Sutsilvan} & \textbf{Surmiran} & \textbf{Puter} & \textbf{Vallader} \\
\midrule
\mbox{MADLAD-400 (10.7B)}      & & & & & & \\
\mbox{– direct} & 58.3 / 63.0 & 52.9 / 54.7 & 40.6 / 38.1 & 45.2 / 40.4 & 49.7 / 49.8 & 52.8 / 52.9 \\
\mbox{– pivoting via English} & 56.1 / 64.9 & 50.3 / 52.8 & 39.4 / 37.4 & 42.4 / 40.1 & 47.2 / 49.8 & 49.3 / 51.9 \\
\midrule
Supertext     & 72.3 / 92.6 & 66.9 / 90.7 & 58.7 / 76.6 & 62.9 / 81.5 & 67.0 / 85.2 & 69.1 / 86.6 \\
\midrule
\mbox{Llama 3.3 (70B)}        & 63.1 / 82.8 & 57.0 / 75.5 & 48.7 / 59.2 & 52.1 / 64.3 & 57.1 / 73.2 & 60.0 / 75.4 \\
GPT-4o        & 74.3 / 92.9 & 70.9 / 92.2 & 64.2 / 85.2 & 67.7 / 87.3 & 71.7 / 90.6 & 75.1 / 91.1 \\
\mbox{Gemini 2.5 Flash}        & \textbf{75.4} / \textbf{93.1} & \textbf{72.1} / \textbf{92.9} & \textbf{68.5} / \textbf{89.4} & \textbf{71.7} / \textbf{90.6} & \textbf{73.5} / \textbf{91.7} & \textbf{77.7} / \textbf{92.3} \\
\bottomrule
\end{tabularx}

\caption{\textbf{Romansh as source language:} ChrF / xCOMET scores of MT systems and LLMs for translation into German from six varieties of Romansh.}\label{tab:romansh-source}
\end{table*}

\begin{table*}[ht]
\centering

\begin{tabularx}{\textwidth}{@{}Xrrrrrr@{}}
\toprule
\textbf{System} & \textbf{Rumantsch Grischun} & \textbf{Sursilvan} & \textbf{Sutsilvan} & \textbf{Surmiran} & \textbf{Puter} & \textbf{Vallader} \\
\midrule
\mbox{MADLAD-400 (10.7B)}      & & & & & & \\
\mbox{– direct} & \phantom{0.00 / }48.0 & \textcolor{graycolor}{\phantom{0.00 / }40.7} & \textcolor{graycolor}{\phantom{0.00 / }34.6} & \textcolor{graycolor}{\phantom{0.00 / }37.1} & \textcolor{graycolor}{\phantom{0.00 / }37.0} & \textcolor{graycolor}{\phantom{0.00 / }38.5} \\
\mbox{– pivoting via English} & \phantom{0.00 / }50.7 & \textcolor{graycolor}{\phantom{0.00 / }43.0} & \textcolor{graycolor}{\phantom{0.00 / }36.1} & \textcolor{graycolor}{\phantom{0.00 / }38.7} & \textcolor{graycolor}{\phantom{0.00 / }38.6} & \textcolor{graycolor}{\phantom{0.00 / }40.1} \\
\midrule
\mbox{Translatur-ia}     & \phantom{0.00 / }19.7 & \textcolor{graycolor}{\phantom{0.00 / }18.1} & \textcolor{graycolor}{\phantom{0.00 / }16.7} & \textcolor{graycolor}{\phantom{0.00 / }17.4} & \textcolor{graycolor}{\phantom{0.00 / }17.3} & \textcolor{graycolor}{\phantom{0.00 / }17.6} \\
Supertext     & \phantom{0.00 / }\textbf{68.9} & \textcolor{graycolor}{\phantom{0.00 / }53.2} & \textcolor{graycolor}{\phantom{0.00 / }43.5} & \textcolor{graycolor}{\phantom{0.00 / }47.8} & \textcolor{graycolor}{\phantom{0.00 / }46.7} & \textcolor{graycolor}{\phantom{0.00 / }49.0} \\
\midrule
\mbox{Llama 3.3 (70B)}        & \phantom{0.00 / }52.1 & \phantom{0.00 / }43.9 & \phantom{0.00 / }36.6 & \phantom{0.00 / }39.3 & \phantom{0.00 / }40.3 & \phantom{0.00 / }42.6 \\
GPT-4o        & \phantom{0.00 / }64.8 & \phantom{0.00 / }\textbf{60.1} & \phantom{0.00 / }41.4 & \phantom{0.00 / }46.4 & \phantom{0.00 / }52.3 & \phantom{0.00 / }55.9 \\
\mbox{Gemini 2.5 Flash}        & \phantom{0.00 / }66.0 & \phantom{0.00 / }58.7 & \phantom{0.00 / }\textbf{43.7} & \phantom{0.00 / }\textbf{50.1} & \phantom{0.00 / }\textbf{53.8} & \phantom{0.00 / }\textbf{57.2} \\
\bottomrule
\end{tabularx}

\caption{\textbf{Romansh as target language:} ChrF scores of MT systems and LLMs for translation from German into Romansh. Results in \textcolor{graycolor}{gray} are based on translations into Rumantsch Grischun, which is the only target variety officially supported by these systems.
}\label{tab:romansh-target}
\end{table*}

%\paragraph{Evaluated Systems}
We use our benchmark to evaluate the following machine translation systems and LLMs:
\begin{itemize}[topsep=0.4em, itemsep=0em]
    \item \textbf{MADLAD-400}~\cite{NEURIPS2023_d49042a5}, a family of open-source MT models trained on parallel data in more than 450 languages, including Romansh. We report results for the largest, 10.7B-parameter model, using sentence segmentation with SpaCy to translate sentences individually, with a beam size of~5.
  \item \textbf{Supertext}, a commercial MT system that supports German and Romansh, among other languages.\footnote{\url{https://supertext.com/}}
   We use the website of Supertext to translate the segments in an Excel file.
  \item \textbf{Translatur-ia}, a closed, early prototype of an MT system that translates from German into Rumantsch Grischun.\footnote{\url{https://translaturia.fhgr.ch/}}
  \item \textbf{Llama 3.3}~\cite{grattafiori2024llama3herdmodels}, an open-source LLM released in November 2024. We use the 70B-parameter version.
  \item \textbf{GPT-4o}~\cite{openai2024gpt4ocard}, a commercial LLM that was released in May 2024 and was billed at \$2.50 per million input tokens and \$10 per million output tokens.%\footnote{\url{https://openai.com/api/pricing/}}
%  \item \textbf{Claude Sonnet 4}~\cite{anthropic2025claude4}, a commercial LLM that was released in May 2025 and is billed at \$3 per million input tokens and \$15 per million output tokens.\footnote{\url{https://docs.anthropic.com/en/docs/about-claude/models/overview}}
  \item \textbf{Gemini 2.5 Flash}~\cite{comanici2025gemini25pushingfrontier}, a commercial LLM that was released in June 2025 and was billed at \$0.30 per million input tokens and \$2.50 per million output tokens. %\footnote{\url{https://cloud.google.com/vertex-ai/generative-ai/pricing}}
   We turn off the `thinking' mode to enable a direct comparison with the other systems.
\end{itemize}

\paragraph{LLM Prompting}
When using LLMs for translation, we use the same prompting setup as the WMT24 General Machine Translation Shared Task~\cite{kocmi-etal-2024-findings}.\footnote{\url{https://github.com/wmt-conference/wmt-collect-translations}}
Specifically, we use 3-shot prompting with temperature set to zero.
The prompt template is listed in Appendix~\ref{sec:prompt}.\footnote{A limitation of this prompt template is that it does not provide the LLM with context beyond the segment that is currently being translated. We opt to keep the setup similar to WMT24 and leave document-level evaluation to future work.}
As few-shot examples, we use typical example sentences from the fable \textit{The Fox and the Crow}~\cite{gross_romansh_2004}, which we list in Appendix~\ref{sec:few-shot-examples}.

\paragraph{Quality Metrics}
For evaluating translation quality, we use ChrF~\cite{popovic-2015-chrf}, a metric based on character n-grams that does not require word segmentation, via SacreBLEU~\cite{post-2018-call}.\footnote{Signature: \\ \texttt{\#:1|c:mixed|e:yes|nc:6|nw:0|s:no|v:2.5.1}}
% While ChrF is a surface-level metric that does not quite achieve the same accuracy as neural or LLM-based metrics~\cite{freitag-etal-2024-llms}, it has the advantage that it is language-agnostic and has been extensively validated as an evaluation metric.

For evaluating translations from Romansh into German, we additionally use xCOMET~\cite{guerreiro-etal-2024-xcomet}, a neural metric that was ranked highly in the WMT24 Metrics Shared Task~\cite{freitag-etal-2024-llms}.
We use model version \texttt{XCOMET-XL}\footnote{\url{https://hf.co/Unbabel/XCOMET-XL}} in the \textit{reference-only} mode, i.e., we do not provide the Romansh source sequence to the metric, a language it has not seen during training.
While xCOMET is designed to support such monolingual, \textit{reference-only} evaluation, this mode has not been as extensively validated as other modes.
Thus, xCOMET complements ChrF but should be interpreted with some caution.
Following~\citet{kocmi-etal-2024-findings}, we report xCOMET as the macro-average over domains to control for different segment granularities.

\section{Results}

\begin{figure}
\centering
\includegraphics{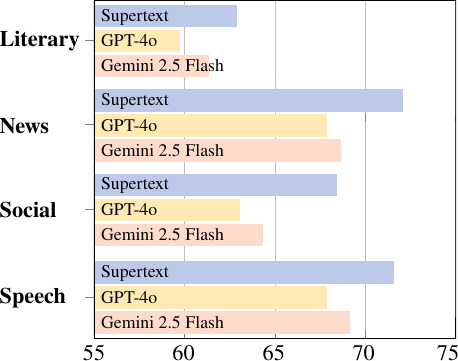}
\caption{Domain-specific ChrF scores of systems translating from German into Rumantsch Grischun.}
\label{fig:chrf-domain}
\end{figure}

\subsection{Comparison of Translation Directions}
Table~\ref{tab:romansh-source} shows the results for translation from Romansh into German, while Table~\ref{tab:romansh-target} shows the results for translation from German into Romansh.
The former performs consistently better than the latter; this is observed both for the supervised MT systems (MADLAD, Supertext) and for the LLMs.

Comparing performance across the six varieties of Romansh, we find that translation out of Romansh into German is relatively robust to linguistic variation: For Gemini, the gap in terms of ChrF between the minimum and maximum is $77.7 - 68.5$.
In contrast, for translation into Romansh, the gap is $66.0 - 43.7$.
Future work could exploit this asymmetry by using back-translation~\cite{sennrich-etal-2016-improving} for augmenting monolingual Romansh text with synthetic German translations.

\subsection{Ranking of Models}
For translation from Romansh into German, we report both ChrF and xCOMET scores in Table~\ref{tab:romansh-source}.
We find that the system rankings are largely consistent between the two metrics, on average over the four domains, with Gemini 2.5 Flash achieving the highest scores according to both metrics.
MADLAD-400 underperforms the other systems, likely due to the limited Romansh training data and the massively multilingual nature of the model.

In the German–Romansh direction, where ChrF is the only available metric~(Table~\ref{tab:romansh-target}), we find that Gemini 2.5 Flash again achieves the highest scores for four out of six varieties.
Supertext is the highest-ranked system for translation into Rumantsch Grischun, which is the officially supported target variety of this product.

\subsection{Domain Difficulty}
Figure~\ref{fig:chrf-domain} compares ChrF scores for German--Rumantsch Grischun translation for the four domains covered by the WMT24++ benchmark.
The figure indicates that the \textit{News} domain is the least challenging for all systems, which is consistent with findings of the WMT24 task for other languages~\cite{kocmi-etal-2024-findings}.
Surprisingly, lowest ChrF scores are achieved in the \textit{Literary} domain, while the human evaluation of the WMT24 task did not find a systematic difference between the \textit{Literary} and \textit{News} domains in terms of difficulty.
The \textit{Speech} domain yields similar scores to \textit{News}, and \textit{Social} is slightly more challenging.
Detailed results for each domain and variety~(Appendices~\ref{app:detailed-rm-de-results} and~\ref{app:detailed-de-rm-results}) indicate that this pattern is consistent across varieties.

\subsection{Target Variety Adherence of LLMs}

\begin{figure}[t]
\centering

\begin{tabularx}{\columnwidth}{@{}X|*{6}{>{\raggedleft\arraybackslash}p{0.65cm}}}
\multirow[b]{2}{*}{$\downarrow$~\textbf{tgt}}
& \multicolumn{6}{l}{$\textbf{ref}$~$\rightarrow$} \\
& RG & Surs.\ & Suts.\ & Surm.\ & Puter & Vall. \\
\midrule
RG     & \cellcolor{uzhblue!50}66.0   & \cellcolor{uzhblue!24}51.6   & \cellcolor{uzhblue!8}42.6   & \cellcolor{uzhblue!16}46.7   & \cellcolor{uzhblue!15}46.2   & \cellcolor{uzhblue!19}48.5   \\
Surs.  & \cellcolor{uzhblue!34}57.0 & \cellcolor{uzhblue!37}58.7 & \cellcolor{uzhblue!8}42.6 & \cellcolor{uzhblue!12}44.8 & \cellcolor{uzhblue!10}43.7 & \cellcolor{uzhblue!13}45.1 \\
Suts.  & \cellcolor{uzhblue!33}56.7 & \cellcolor{uzhblue!22}50.3 & \cellcolor{uzhblue!10}43.7 & \cellcolor{uzhblue!15}46.6 & \cellcolor{uzhblue!11}44.1 & \cellcolor{uzhblue!14}45.8 \\
Surm.  & \cellcolor{uzhblue!26}52.3 & \cellcolor{uzhblue!15}46.5 & \cellcolor{uzhblue!8}42.5 & \cellcolor{uzhblue!22}50.1 & \cellcolor{uzhblue!10}43.8 & \cellcolor{uzhblue!12}44.5 \\
Puter  & \cellcolor{uzhblue!17}47.5   & \cellcolor{uzhblue!7}42.1   & \cellcolor{uzhblue!0}38.0   & \cellcolor{uzhblue!6}41.2   & \cellcolor{uzhblue!28}53.8   & \cellcolor{uzhblue!30}54.6   \\
Vall.  & \cellcolor{uzhblue!21}49.9 & \cellcolor{uzhblue!9}43.2 & \cellcolor{uzhblue!1}38.6 & \cellcolor{uzhblue!7}41.9 & \cellcolor{uzhblue!26}52.7 & \cellcolor{uzhblue!34}57.2 \\
\bottomrule
\end{tabularx}

\caption{``Confusion matrix'' of Gemini 2.5 Flash when translating into specific Romansh varieties. To visualize the degree to which the LLM output matches the requested variety (\textbf{tgt}), we evaluate the outputs with each set of reference translations (\textbf{ref}). A system that adheres to the requested target variety will achieve higher ChrF scores in the diagonal cells than in the off-diagonal cells.}
\label{tab:llm-confusion-matrix}
\end{figure}

While MADLAD and Supertext are limited to Rumantsch Grischun as the target variety, the LLMs can be prompted to produce translations in any of the six varieties.
This raises the question of whether the LLMs actually adhere to the requested target variety.
%The results in Table~\ref{tab:romansh-target} confirm that state-of-the-art LLMs can achieve some standard of translation quality for all the idioms (e.g., higher than or competitive to the scores achieved by the human-written reference translations for the other idioms, shown in Table~\ref{tab:pairwise-chrf}), but these results do not directly quantify adherence to the target variety.
Figure~\ref{tab:llm-confusion-matrix} shows a ``confusion matrix'' for Gemini 2.5 Flash, where we evaluate the system output not only with the reference translations for the requested target variety, but also with contrastive reference translations for the other varieties.
%Our expectation is that a system that adheres to the requested target variety will achieve the highest ChrF scores in the diagonal cells, where the target variety is equal to the reference variety.
The results suggest that state-of-the-art LLMs already have some degree of idiom awareness, but gravitate towards the higher-resource varieties (Rumantsch Grischun, Sursilvan, and Vallader).

\section{Conclusion}
This work fills a long-standing gap in the evaluation of machine translation for the Romansh language: the creation of a benchmark for the six main varieties of Romansh, and the provision of baseline results for existing MT systems and LLMs that cover Romansh.
%We hope that our benchmark and baseline results will help to foster further research on idiom-aware MT for Romansh.
% Future work: Dialects?

\section*{Acknowledgements}
We thank RTR and Fundaziun Patrimoni Cultural RTR for their support.
We are grateful to Kirill Semenov for assistance with the poster presentation, Zachary Hopton for assistance with editing the manuscript, Noëmi Aepli and Martin Volk for helpful advice, and Tom Kocmi for help with the WMT24 benchmark.

Figure~\ref{fig:idiom-map} uses maps from the \href{https://www.gr.ch/DE/institutionen/verwaltung/dfg/afg/gemeinden/UnsereGemeinden/Seiten/Einteilung.aspx}{Canton of Grisons} and from Wikimedia Commons (\href{https://commons.wikimedia.org/wiki/File:Switzerland_administrative_divisions_-_colored_unlabelled.svg}{User:Tomchen1989, User:NordNordWest, User:TUBS},
\href{https://creativecommons.org/licenses/by-sa/3.0}{CC BY-SA 3.0}).

\section*{Author Contributions}
JV: Conceptualization, Funding acquisition, Investigation, Methodology, Project administration, Supervision, Writing – original draft. \\
IPP: Funding acquisition, Project administration, Supervision, Writing – review \& editing. \\
NBS: Writing – original draft. \\
SBG, AB, SB, MC, GPG, FH, GH, AL, VL, WR: Translation and/or translation quality assurance. \\
LD, BV: Linguistic supervision of translation workflow. \\
AR: Software. \\
RS: Funding acquisition, Methodology, Project administration, Supervision, Writing – review \& editing. \\

% Bibliography entries for the entire Anthology, followed by custom entries
%\bibliography{anthology,custom}
% Custom bibliography entries only
\bibliography{bibliography}

\appendix

\clearpage

\onecolumn

\section{Dataset Statistics}
\label{sec:dataset-stats}

\begin{table}[h]
\centering
\begin{tabularx}{\textwidth}{@{}Xrrrrrrrrrr@{}}
\toprule
\textbf{Variety} & \multicolumn{5}{c}{\textbf{Segments}} & \multicolumn{5}{c}{\textbf{Tokens}} \\
\cmidrule(lr){2-6}\cmidrule(lr){7-11}
 & \textbf{Lit.} & \textbf{News} & \textbf{Soc.} & \textbf{Speech} & \textbf{Total} & \textbf{Lit.} & \textbf{News} & \textbf{Soc.} & \textbf{Speech} & \textbf{Total} \\
\midrule
German \mbox{\cite{deutsch2025wmt24expandinglanguagecoverage}} & 206 & 149 & 531 & 111 & 998 & 9\,814 & 9\,416 & 10\,556 & 10\,157 & 39\,954 \\
\midrule
RG & 206 & 149 & 531 & 111 & 998 & 10\,893 & 11\,626 & 11\,949 & 10\,174 & 44\,653 \\
Sursilvan & 206 & 149 & 531 & 111 & 998 & 10\,609 & 11\,287 & 11\,913 & 10\,144 & 43\,964 \\
Sutsilvan & 206 & 149 & 531 & 111 & 998 & 11\,147 & 11\,667 & 12\,241 & 10\,378 & 45\,444 \\
Surmiran & 206 & 149 & 531 & 111 & 998 & 10\,718 & 11\,424 & 11\,907 & 10\,167 & 44\,227 \\
Puter & 206 & 149 & 531 & 111 & 998 & 11\,325 & 11\,595 & 12\,267 & 10\,490 & 45\,688 \\
Vallader & 206 & 149 & 531 & 111 & 998 & 11\,412 & 11\,700 & 12\,330 & 10\,513 & 45\,966 \\
\bottomrule
\end{tabularx}
\caption{Dataset statistics for each language variety. Token counts are based on whitespace tokenization.}
\label{tab:dataset-stats}
\end{table}

\bigskip

\section{Prompt Template for LLMs}
\label{sec:prompt}

The below example is parameterized as follows:
\begin{itemize}[topsep=0.4em, itemsep=0em]
    \item Source language: German
    \item Target language: Romansh
    \item Target variety: Rumantsch Grischun
    \item Source sentence: \textit{``Sisos Darstellungen von Land und Wasser in neuer Ausstellung''}
\end{itemize}

\begin{tcolorbox}[colback=white,colframe=black!75!black,boxrule=0.7pt,arc=2pt,left=2pt,right=2pt,top=2pt,bottom=2pt]
\noindent{}\texttt{Translate the following segment surrounded in triple backticks into Romansh (Rumantsch Grischun variety). The German segment:} \\

\noindent{}\texttt{\textasciigrave\textasciigrave\textasciigrave Sisos Darstellungen von Land und Wasser in neuer Ausstellung\textasciigrave\textasciigrave\textasciigrave}
\end{tcolorbox}

\clearpage

\section{Few-shot Examples for LLM Prompting}
\label{sec:few-shot-examples}

\begin{description}[leftmargin=!,labelwidth=2.5em,itemsep=0.4em,parsep=0em,topsep=0em]
\item[German:]
\item Der Fuchs war wieder einmal hungrig.
\item Da sah er auf einer Tanne einen Raben, der ein Stück Käse in seinem Schnabel hielt.
\item Das würde mir schmecken, dachte er, und rief dem Raben zu: «Wie schön du bist! Wenn dein Gesang ebenso schön ist wie dein Aussehen, dann bist du der Schönste von allen Vögeln».

\item[English:]
\item The fox was hungry yet again.
\item There, he saw a raven upon a fir holding a piece of cheese in its beak.
\item This I would like, he thought, and shouted at the raven: "You are so beautiful! If your singing is as beautiful as your looks, then you are the most beautiful of all birds."

\item[Rumantsch Grischun:]
\item La vulp era puspè ina giada fomentada.
\item Qua ha ella vis sin in pign in corv che tegneva in toc chaschiel en ses pichel.
\item Quai ma gustass, ha ella pensà, ed ha clamà al corv: «Tge bel che ti es! Sche tes chant è uschè bel sco tia parita, lur es ti il pli bel utschè da tuts».

\item[Sursilvan:]
\item L'uolp era puspei inagada fomentada.
\item Cheu ha ella viu sin in pegn in tgaper che teneva in toc caschiel en siu bec.
\item Quei gustass a mi, ha ella tertgau, ed ha clamau al tgaper: «Tgei bi che ti eis! Sche tiu cant ei aschi bials sco tia cumparsa, lu eis ti il pli bi utschi da tuts».

\item[Sutsilvan:]
\item La gualp eara puspe egn'eada fumantada.
\item Qua â ella vieu sen egn pegn egn corv ca taneva egn toc caschiel ainten sieus pecel.
\item Quegl gustass a mei, â ella tartgieu, ed ha clamo agli corv: «Tge beal ca tei es! Scha tieus tgànt e aschi beal sco tia pareta, alura es tei igl ple beal utschi da tuts».

\item[Surmiran:]
\item La golp era puspe eneda famantada.
\item Co ò ella via sen en pegn en corv tgi tigniva en toc caschiel an sies pecal.
\item Chegl am gustess, ò ella panso, ed ò clamo agl corv: «Tge bel tgi te ist! Schi ties cant è schi bel scu tia parentscha, alloura ist te igl pi bel utschel da tots».

\item[Puter:]
\item La vuolp d'eira darcho üna vouta famanteda.
\item Co ho'la vis sün ün pin ün corv chi tgnaiva ün töch chaschöl in sieu pical.
\item Que am gustess, ho'la penso, ed ho clamo al corv: «Che bel cha tü est! Scha tieu chaunt es uschè bel scu tia apparentscha, alura est tü il pü bel utschè da tuots».

\item[Vallader:]
\item La vuolp d'eira darcheu üna jada fomantada.
\item Qua ha'la vis sün ün pin ün corv chi tgnaiva ün toc chaschöl in seis pical.
\item Quai am gustess, ha'la pensà, ed ha clomà al corv: «Che bel cha tü est! Scha teis chant es uschè bel sco tia apparentscha, lura est tü il plü bel utschè da tuots».
\end{description}

\clearpage

\section{Translation Guidelines (in German)}
\label{sec:guidelines}

\subsection*{1. Wofür werden die Übersetzungen benötigt?}

Wir verwenden die von Ihnen erstellen Übersetzungen für die Evaluierung von maschinellen Übersetzungssystemen. Der Output der Übersetzungssysteme wird mit Ihrer Übersetzung verglichen – je ähnlicher der Output, desto besser das Übersetzungssystem.

Die Referenzübersetzungen werden nicht für das Training der Übersetzungssysteme verwendet.

\subsection*{2. Wie wurden die Texte ausgewählt?}

Die Texte stammen aus einem bestehenden Datensatz («WMT24»), der zuvor bereits aus dem Englischen in 55 verschiedene Sprachen übersetzt worden ist, darunter Deutsch.

Der Datensatz setzt sich aus vier Textsorten zusammen:

\begin{itemize}
    \item \textbf{literary:} Fan-Fiction, welche auf der Website «Archive of Our Own» veröffentlicht wurde.
    \item \textbf{news:} Zufällig ausgewählte Online-News vom Januar 2024.
    \item \textbf{social:} Zufällig ausgewählte Threads aus dem Sozialen Netzwerk «Mastodon».
    \item \textbf{speech:} Transkripte zufällig ausgewählter YouTube-Videos.
\end{itemize}

\subsection*{3. Wie ist die Excel-Datei aufgebaut?}

Wir erstellen für jedes Idiom eine eigene Excel-Datei. Die Datei enthält vier Tabellen für die vier Textsorten, und jede Zeile in der Tabelle entspricht einem Textsegment. Mehrere Textsegmente setzen sich zu zusammenhängenden Dokumenten zusammen, welche durch graue Leerzeilen voneinander abgetrennt sind.

Die Tabellen sind wie folgt aufgebaut:

\begin{itemize}
    \item \textbf{English:} Originaler Text auf Englisch.
    \item \textbf{document\_id:} ID des Dokuments.
    \item \textbf{segment\_id:} ID des Segments.
    \item \textbf{url:} Webseite, von welcher das Dokument ursprünglich bezogen wurde. Kann optional aufgerufen werden, um den Kontext des Dokuments nachzuvollziehen.
    \item \textbf{German:} Referenzübersetzung auf Deutsch.
    \item \textbf{translation:} Hier soll die Übersetzung auf Rätoromanisch eingetragen werden.
    \item \textbf{comment:} Kann von Ihnen optional benutzt werden, um uns einen wichtigen Kommentar zu hinterlassen.
\end{itemize}

\subsection*{4. Welche Anforderungen gelten an die Übersetzungen?}

\begin{itemize}
    \item Die Referenzübersetzungen repräsentieren Ihre Erwartungen an den Output eines guten maschinellen Übersetzungssystems. Überlegen Sie sich: Was für eine Übersetzung würde ein gutes Übersetzungssystem (im Stil von DeepL oder Google Translate) für das entsprechende romanische Idiom erzeugen?
    \item Massgeblich für Ihre Übersetzung sollte der Text auf Deutsch sein. Der englische Text kann optional zu Rate gezogen werden, falls der deutsche Text mehrdeutig ist.
    \item Weil wir Ihre Referenzübersetzungen für die Evaluierung verwenden möchten, bitten wir Sie, keine AI-Tools zu verwenden. Bitte erstellen Sie die Übersetzungen von Grund auf in Ihren eigenen Worten.
    \begin{itemize}
        \item Erlaubte Hilfsmittel: Wörterbücher, Translation Memories, \ldots
        \item Nicht erlaubte Hilfsmittel: ChatGPT, Copilot, Supertext/Textshuttle, DeepL, \ldots
    \end{itemize}
    Wir können den Übersetzungen ansehen, wenn sie mit AI-Unterstützung erstellt wurden, und wir müssten im schlimmsten Fall die Übersetzungen noch einmal neu erstellen lassen (dies ist leider in der Vergangenheit schon vorgekommen).
    \item Bitte fügen Sie keine Übersetzungsalternativen oder Erklärungen in Klammern in die Übersetzungen ein.
    \item Bitte übersetzen Sie die Information in den Texten vollständig.
    \item Lehnwörter, Eigennamen etc. dürfen gerne auf Englisch oder Deutsch belassen werden, falls ein gutes rätoromanisches Übersetzungssystem das gleiche machen sollte.
    \item Betreffend die Textsorte «social»:
    \begin{itemize}
        \item Viele Posts enthalten Fachbegriffe, Slang, Abkürzungen. Durch die Übersetzung vom Englischen ins Deutsche dürfte vieles schon weniger kryptisch geworden sein. Falls dennoch bei einem Post Unsicherheit besteht, können Ihnen diese Notizen weiterhelfen: \url{https://github.com/wmt-conference/wmt24-news-systems/blob/main/README-social-domain-translation-notes.pdf}
        \item Nutzernamen wurden anonymisiert (z.B. @user1, @user2). Bitte übernehmen Sie die Nutzernamen eins zu eins, d.h. Sie müssen diese nicht übersetzen.
        \item Hingegen dürfen \#Hashtags gerne übersetzt werden, falls dies im Kontext Sinn macht.
    \end{itemize}
\end{itemize}

\clearpage

\section{Detailed Results for RM--DE}
\label{app:detailed-rm-de-results}
\subsection{Rumantsch Grischun to German}

\begin{table}[H]
\centering
\begin{tabularx}{\textwidth}{@{}Xrrrrr@{}}
\toprule
\textbf{System} & \textbf{Literary} & \textbf{News} & \textbf{Social} & \textbf{Speech} & \textbf{Macro-Average} \\
\midrule
\mbox{MADLAD-400 10.7B}      & & & & & \\
\mbox{– direct} & 50.6 / 52.8 & 67.9 / 81.3 & 53.0 / 71.7 & 59.5 / 46.4 & 57.8 / 63.0 \\
\mbox{– pivoting via English} & 49.8 / 57.1 & 63.2 / 79.1 & 51.9 / 73.1 & 58.0 / 50.2 & 55.7 / 64.9 \\
\midrule
Supertext     & 68.3 / 89.9 & 71.1 / 95.1 & 73.9 / \textbf{96.2} & 75.8 / 89.1 & 72.3 / 92.6 \\
\midrule
Llama 3.3 (70B)        & 56.8 / 75.3 & 68.0 / 91.3 & 58.2 / 84.3 & 68.6 / 80.4 & 62.9 / 82.8 \\
GPT-4o        & \textbf{71.5} / \textbf{91.8} & 74.1 / \textbf{95.9} & 73.6 / 94.3 & 78.3 / 89.6 & 74.4 / 92.9 \\
Gemini 2.5 Flash        & \textbf{71.5} / 90.6 & \textbf{74.4} / 95.7 & \textbf{76.5} / 95.9 & \textbf{79.5} / \textbf{90.2} & \textbf{75.5} / \textbf{93.1} \\
\bottomrule
\end{tabularx}
\caption{ChrF / xCOMET scores for translation from Rumantsch Grischun into German.}
\end{table}

\subsection{Sursilvan to German}

\begin{table}[H]
\centering
\begin{tabularx}{\textwidth}{@{}Xrrrrr@{}}
\toprule
\textbf{System} & \textbf{Literary} & \textbf{News} & \textbf{Social} & \textbf{Speech} & \textbf{Macro-Average} \\
\midrule
\mbox{MADLAD-400 10.7B}      & & & & & \\
\mbox{– direct} & 45.7 / 44.0 & 64.4 / 76.6 & 46.9 / 65.8 & 52.2 / 32.5 & 52.3 / 54.7 \\
\mbox{– pivoting via English} & 44.0 / 45.0 & 59.1 / 70.2 & 45.2 / 65.8 & 51.1 / 30.4 & 49.9 / 52.8 \\
\midrule
Supertext     & 63.6 / 87.1 & 68.7 / 94.8 & 64.5 / 94.4 & 70.6 / 86.4 & 66.9 / 90.7 \\
\midrule
Llama 3.3 (70B)        & 50.5 / 63.7 & 64.3 / 88.4 & 50.4 / 77.0 & 61.7 / 72.9 & 56.7 / 75.5 \\
GPT-4o        & 68.6 / \textbf{90.7} & 72.2 / 95.5 & 67.8 / 93.0 & \textbf{74.9} / \textbf{89.5} & 70.9 / 92.2 \\
Gemini 2.5 Flash        & \textbf{69.7} / 90.5 & \textbf{72.9} / \textbf{95.7} & \textbf{70.8} / \textbf{95.7} & \textbf{74.9} / \textbf{89.5} & \textbf{72.1} / \textbf{92.9} \\
\bottomrule
\end{tabularx}
\caption{ChrF / xCOMET scores for translation from Sursilvan into German.}
\end{table}

\subsection{Sutsilvan to German}

\begin{table}[H]
\centering
\begin{tabularx}{\textwidth}{@{}Xrrrrr@{}}
\toprule
\textbf{System} & \textbf{Literary} & \textbf{News} & \textbf{Social} & \textbf{Speech} & \textbf{Macro-Average} \\
\midrule
\mbox{MADLAD-400 10.7B}      & & & & & \\
\mbox{– direct} & 33.1 / 31.4 & 54.1 / 50.3 & 32.8 / 48.4 & 40.0 / 22.3 & 40.0 / 38.1 \\
\mbox{– pivoting via English} & 32.9 / 31.5 & 50.3 / 47.2 & 33.1 / 49.8 & 39.4 / 21.2 & 38.9 / 37.4 \\
\midrule
Supertext     & 55.1 / 67.9 & 62.9 / 87.8 & 54.7 / 86.1 & 61.1 / 64.8 & 58.5 / 76.6 \\
\midrule
Llama 3.3 (70B)        & 43.0 / 44.3 & 57.6 / 74.8 & 40.3 / 61.8 & 52.9 / 55.7 & 48.4 / 59.2 \\
GPT-4o        & 61.5 / 80.5 & 70.0 / 93.3 & 54.4 / 81.6 & 70.3 / 85.2 & 64.1 / 85.2 \\
Gemini 2.5 Flash        & \textbf{65.7} / \textbf{85.0} & \textbf{71.2} / \textbf{94.2} & \textbf{64.6} / \textbf{92.4} & \textbf{72.3} / \textbf{86.1} & \textbf{68.5} / \textbf{89.4} \\
\bottomrule
\end{tabularx}
\caption{ChrF / xCOMET scores for translation from Sutsilvan into German.}
\end{table}

\vfill
\clearpage
\subsection{Surmiran to German}

\begin{table}[H]
\centering
\begin{tabularx}{\textwidth}{@{}Xrrrrr@{}}
\toprule
\textbf{System} & \textbf{Literary} & \textbf{News} & \textbf{Social} & \textbf{Speech} & \textbf{Macro-Average} \\
\midrule
\mbox{MADLAD-400 10.7B}      & & & & & \\
\mbox{– direct} & 37.8 / 34.1 & 58.6 / 53.0 & 37.9 / 52.7 & 43.9 / 21.7 & 44.5 / 40.4 \\
\mbox{– pivoting via English} & 36.9 / 36.1 & 53.0 / 50.4 & 36.0 / 50.6 & 42.0 / 23.1 & 41.9 / 40.1 \\
\midrule
Supertext     & 59.6 / 76.9 & 67.8 / 91.6 & 58.9 / 88.6 & 64.4 / 69.0 & 62.7 / 81.5 \\
\midrule
Llama 3.3 (70B)        & 46.7 / 52.9 & 60.9 / 78.7 & 44.2 / 66.0 & 55.2 / 59.7 & 51.7 / 64.3 \\
GPT-4o        & 66.4 / 86.2 & 71.6 / 92.3 & 61.3 / 85.6 & 71.5 / 85.1 & 67.7 / 87.3 \\
Gemini 2.5 Flash        & \textbf{69.3} / \textbf{88.5} & \textbf{72.8} / \textbf{94.3} & \textbf{69.6} / \textbf{93.0} & \textbf{75.2} / \textbf{86.7} & \textbf{71.7} / \textbf{90.6} \\
\bottomrule
\end{tabularx}
\caption{ChrF / xCOMET scores for translation from Surmiran into German.}
\end{table}

\subsection{Puter to German}

\begin{table}[H]
\centering
\begin{tabularx}{\textwidth}{@{}Xrrrrr@{}}
\toprule
\textbf{System} & \textbf{Literary} & \textbf{News} & \textbf{Social} & \textbf{Speech} & \textbf{Macro-Average} \\
\midrule
\mbox{MADLAD-400 10.7B}      & & & & & \\
\mbox{– direct} & 42.6 / 41.0 & 63.3 / 68.7 & 41.8 / 58.6 & 48.2 / 30.9 & 49.0 / 49.8 \\
\mbox{– pivoting via English} & 41.9 / 42.9 & 57.1 / 63.2 & 40.4 / 60.2 & 47.4 / 32.9 & 46.7 / 49.8 \\
\midrule
Supertext     & 63.9 / 81.6 & 71.3 / 93.4 & 62.6 / 89.4 & 69.5 / 76.4 & 66.8 / 85.2 \\
\midrule
Llama 3.3 (70B)        & 52.5 / 63.6 & 64.6 / 84.8 & 49.0 / 73.8 & 61.0 / 70.4 & 56.8 / 73.2 \\
GPT-4o        & 69.7 / 88.7 & 73.6 / 94.0 & 67.0 / 91.7 & 76.6 / \textbf{88.0} & 71.7 / 90.6 \\
Gemini 2.5 Flash        & \textbf{71.5} / \textbf{89.7} & \textbf{74.6} / \textbf{94.6} & \textbf{70.7} / \textbf{94.5} & \textbf{77.1} / \textbf{88.0} & \textbf{73.5} / \textbf{91.7} \\
\bottomrule
\end{tabularx}
\caption{ChrF / xCOMET scores for translation from Puter into German.}
\end{table}

\subsection{Vallader to German}

\begin{table}[H]
\centering
\begin{tabularx}{\textwidth}{@{}Xrrrrr@{}}
\toprule
\textbf{System} & \textbf{Literary} & \textbf{News} & \textbf{Social} & \textbf{Speech} & \textbf{Macro-Average} \\
\midrule
\mbox{MADLAD-400 10.7B}      & & & & & \\
\mbox{– direct} & 45.9 / 46.9 & 65.0 / 71.5 & 46.1 / 63.8 & 51.7 / 29.3 & 52.2 / 52.9 \\
\mbox{– pivoting via English} & 43.1 / 47.8 & 59.6 / 65.7 & 43.7 / 64.3 & 48.7 / 29.7 & 48.8 / 51.9 \\
\midrule
Supertext     & 65.6 / 83.3 & 71.6 / 92.5 & 66.7 / 91.7 & 72.1 / 78.9 & 69.0 / 86.6 \\
\midrule
Llama 3.3 (70B)        & 54.7 / 67.2 & 67.3 / 86.2 & 52.8 / 78.6 & 64.1 / 69.5 & 59.8 / 75.4 \\
GPT-4o        & 72.4 / 90.1 & 77.3 / 94.5 & 71.2 / 91.2 & 79.4 / \textbf{88.4} & 75.1 / 91.1 \\
Gemini 2.5 Flash        & \textbf{74.7} / \textbf{91.1} & \textbf{78.9} / \textbf{94.7} & \textbf{76.1} / \textbf{95.3} & \textbf{80.9} / 88.1 & \textbf{77.6} / \textbf{92.3} \\
\bottomrule
\end{tabularx}
\caption{ChrF / xCOMET scores for translation from Vallader into German.}
\end{table}

\vfill
\clearpage

\clearpage

\section{Detailed Results for DE--RM}
\label{app:detailed-de-rm-results}
\subsection{German to Rumantsch Grischun}

\begin{table}[H]
\centering
\begin{tabularx}{\textwidth}{@{}Xrrrrr@{}}
\toprule
\textbf{System} & \textbf{Literary} & \textbf{News} & \textbf{Social} & \textbf{Speech} & \textbf{Macro-Average} \\
\midrule

\mbox{MADLAD-400 10.7B}      & & & & & \\
\mbox{– direct} & \phantom{0.00 / }42.9 & \phantom{0.00 / }56.3 & \phantom{0.00 / }42.6 & \phantom{0.00 / }48.6 & \phantom{0.00 / }47.6 \\
\mbox{– pivoting via English} & \phantom{0.00 / }45.6 & \phantom{0.00 / }56.6 & \phantom{0.00 / }47.6 & \phantom{0.00 / }51.8 & \phantom{0.00 / }50.4 \\
\midrule
Translatur-ia     & \phantom{0.00 / }17.1 & \phantom{0.00 / }20.1 & \phantom{0.00 / }23.4 & \phantom{0.00 / }17.2 & \phantom{0.00 / }19.5 \\
Supertext     & \phantom{0.00 / }\textbf{62.8} & \phantom{0.00 / }\textbf{72.1} & \phantom{0.00 / }\textbf{68.4} & \phantom{0.00 / }\textbf{71.6} & \phantom{0.00 / }\textbf{68.7} \\
\midrule

Llama 3.3 (70B)        & \phantom{0.00 / }46.8 & \phantom{0.00 / }57.7 & \phantom{0.00 / }48.6 & \phantom{0.00 / }54.0 & \phantom{0.00 / }51.8 \\
GPT-4o        & \phantom{0.00 / }59.9 & \phantom{0.00 / }67.9 & \phantom{0.00 / }63.0 & \phantom{0.00 / }67.9 & \phantom{0.00 / }64.7 \\
Gemini 2.5 Flash        & \phantom{0.00 / }61.3 & \phantom{0.00 / }68.6 & \phantom{0.00 / }64.3 & \phantom{0.00 / }69.2 & \phantom{0.00 / }65.8 \\
\bottomrule
\end{tabularx}
\caption{ChrF scores for translation from German into Rumantsch Grischun.}
\end{table}

\subsection{German to Sursilvan}

\begin{table}[H]
\centering
\begin{tabularx}{\textwidth}{@{}Xrrrrr@{}}
\toprule
\textbf{System} & \textbf{Literary} & \textbf{News} & \textbf{Social} & \textbf{Speech} & \textbf{Macro-Average} \\
\midrule

Llama 3.3 (70B)        & \phantom{0.00 / }39.7 & \phantom{0.00 / }50.7 & \phantom{0.00 / }39.4 & \phantom{0.00 / }44.5 & \phantom{0.00 / }43.6 \\
GPT-4o        & \phantom{0.00 / }\textbf{56.2} & \phantom{0.00 / }\textbf{63.7} & \phantom{0.00 / }\textbf{57.5} & \phantom{0.00 / }\textbf{62.4} & \phantom{0.00 / }\textbf{59.9} \\
Gemini 2.5 Flash        & \phantom{0.00 / }55.4 & \phantom{0.00 / }61.8 & \phantom{0.00 / }56.4 & \phantom{0.00 / }60.6 & \phantom{0.00 / }58.5 \\
\bottomrule
\end{tabularx}
\caption{ChrF scores for translation from German into Sursilvan.}
\end{table}

\subsection{German to Sutsilvan}

\begin{table}[H]
\centering
\begin{tabularx}{\textwidth}{@{}Xrrrrr@{}}
\toprule
\textbf{System} & \textbf{Literary} & \textbf{News} & \textbf{Social} & \textbf{Speech} & \textbf{Macro-Average} \\
\midrule

Llama 3.3 (70B)        & \phantom{0.00 / }32.4 & \phantom{0.00 / }43.0 & \phantom{0.00 / }33.0 & \phantom{0.00 / }36.8 & \phantom{0.00 / }36.3 \\
GPT-4o        & \phantom{0.00 / }37.1 & \phantom{0.00 / }47.5 & \phantom{0.00 / }39.6 & \phantom{0.00 / }40.1 & \phantom{0.00 / }41.1 \\
Gemini 2.5 Flash        & \phantom{0.00 / }\textbf{40.5} & \phantom{0.00 / }48.9 & \phantom{0.00 / }39.8 & \phantom{0.00 / }\textbf{44.5} & \phantom{0.00 / }\textbf{43.4} \\
\bottomrule
\end{tabularx}
\caption{ChrF scores for translation from German into Sutsilvan.}
\end{table}

\subsection{German to Surmiran}

\begin{table}[H]
\centering
\begin{tabularx}{\textwidth}{@{}Xrrrrr@{}}
\toprule
\textbf{System} & \textbf{Literary} & \textbf{News} & \textbf{Social} & \textbf{Speech} & \textbf{Macro-Average} \\
\midrule

Llama 3.3 (70B)        & \phantom{0.00 / }34.7 & \phantom{0.00 / }46.5 & \phantom{0.00 / }35.5 & \phantom{0.00 / }38.8 & \phantom{0.00 / }38.9 \\
GPT-4o        & \phantom{0.00 / }42.2 & \phantom{0.00 / }53.5 & \phantom{0.00 / }43.1 & \phantom{0.00 / }44.8 & \phantom{0.00 / }45.9 \\
Gemini 2.5 Flash        & \phantom{0.00 / }\textbf{46.9} & \phantom{0.00 / }\textbf{55.7} & \phantom{0.00 / }\textbf{47.0} & \phantom{0.00 / }\textbf{49.6} & \phantom{0.00 / }\textbf{49.8} \\
\bottomrule
\end{tabularx}
\caption{ChrF scores for translation from German into Surmiran.}
\end{table}

\subsection{German to Puter}

\begin{table}[H]
\centering
\begin{tabularx}{\textwidth}{@{}Xrrrrr@{}}
\toprule
\textbf{System} & \textbf{Literary} & \textbf{News} & \textbf{Social} & \textbf{Speech} & \textbf{Macro-Average} \\
\midrule

Llama 3.3 (70B)        & \phantom{0.00 / }35.8 & \phantom{0.00 / }47.0 & \phantom{0.00 / }36.0 & \phantom{0.00 / }41.0 & \phantom{0.00 / }39.9 \\
GPT-4o        & \phantom{0.00 / }48.0 & \phantom{0.00 / }57.5 & \phantom{0.00 / }\textbf{49.2} & \phantom{0.00 / }53.4 & \phantom{0.00 / }52.0 \\
Gemini 2.5 Flash        & \phantom{0.00 / }\textbf{51.3} & \phantom{0.00 / }\textbf{59.9} & \phantom{0.00 / }48.5 & \phantom{0.00 / }\textbf{54.8} & \phantom{0.00 / }\textbf{53.6} \\
\bottomrule
\end{tabularx}
\caption{ChrF scores for translation from German into Puter.}
\end{table}

\subsection{German to Vallader}

\begin{table}[H]
\centering
\begin{tabularx}{\textwidth}{@{}Xrrrrr@{}}
\toprule
\textbf{System} & \textbf{Literary} & \textbf{News} & \textbf{Social} & \textbf{Speech} & \textbf{Macro-Average} \\
\midrule

Llama 3.3 (70B)        & \phantom{0.00 / }38.3 & \phantom{0.00 / }49.2 & \phantom{0.00 / }38.6 & \phantom{0.00 / }42.9 & \phantom{0.00 / }42.2 \\
GPT-4o        & \phantom{0.00 / }52.7 & \phantom{0.00 / }60.2 & \phantom{0.00 / }53.2 & \phantom{0.00 / }56.5 & \phantom{0.00 / }55.7 \\
Gemini 2.5 Flash        & \phantom{0.00 / }\textbf{53.8} & \phantom{0.00 / }\textbf{61.6} & \phantom{0.00 / }\textbf{54.2} & \phantom{0.00 / }\textbf{58.2} & \phantom{0.00 / }\textbf{57.0} \\
\bottomrule
\end{tabularx}
\caption{ChrF scores for translation from German into Vallader.}
\end{table}

\clearpage

\section{Examples for System Outputs}
\label{app:system-examples}
\subsection{German to Romansh}
\noindent{}[English:]  \textit{it seems like even iMessage over WiFi isn't working, which doesn't quite make sense to me}\\[0.3em]
\noindent{}German source:  \textit{Anscheinend funktioniert nicht mal iMessage über WiFi, was mir nicht ganz einleuchtet}
\subsubsection*{Rumantsch Grischun}
\noindent{}\underline{Reference:} \textit{Para che gnanc iMessage funcziunia via WiFi, tge ch'è per mai betg dal tut evident.}\\[0.3em]
\noindent{}MADLAD-400 (10.7B) direct:  \textit{Sembla ch'iMessage na funcziunia betg sur WiFi, quai che n'ha nagin sens per mai}\\[0.3em]
\noindent{}Translatur-ia:  \textit{tuttinascha fitg lungascha betg pitschnas la kieni sur la qualafizistgeli , ma prognostitgà}\\[0.3em]
\noindent{}Supertext:  \textit{Para na funcziuna gnanc iMessage sur WiFi, quai che ma resplenda betg dal tut}\\[0.3em]
\noindent{}Llama 3.3 (70B):  \textit{Apparentamain funcziuna schiz ni iMessage sur WiFi, quai ch'ha a mai betg dal tut enclaus.}\\[0.3em]
\noindent{}GPT-4o:  \textit{Apparentamain na funcziuna gnanc iMessage sur WiFi, quai che na ma para betg del tut cler}\\[0.3em]
\noindent{}Gemini 2.5 Flash:  \textit{Apparentamain na funcziunescha gnanc iMessage sur WiFi, quai che na ma para betg dal tuttafatg cler.}\\
\subsubsection*{Sursilvan}
\noindent{}\underline{Reference:} \textit{Sco ei para funcziunescha gnanc iMessage sur WiFi, quei ch'jeu sai buca propi capir.}\\[0.3em]
\noindent{}Llama 3.3 (70B):  \textit{Scheppa che sch'iva nean iMessage sur WiFi, quai che mi n'entuppa betg dal tut}\\[0.3em]
\noindent{}GPT-4o:  \textit{Apparentamein funcziunescha buc ne iMessage sur WiFi, quei cheu ei buc clar a mi}\\[0.3em]
\noindent{}Gemini 2.5 Flash:  \textit{Apparentamein funcziunescha gnanc iMessage sur WiFi, quei che jeu capeschel buca dal tut.}\\
\subsubsection*{Sutsilvan}
\noindent{}\underline{Reference:} \textit{Para funcziunescha gnànc iMessage sur igl WiFi, tge ca fa betga propi sen tanor me}\\[0.3em]
\noindent{}Llama 3.3 (70B):  \textit{Apparentamain funcziuna schi betg iMessage sur WiFi, sche mi nu cumpreta betg del tut}\\[0.3em]
\noindent{}GPT-4o:  \textit{Apparentamein fonzionescha gnanc iMessage sur WiFi, tgei ca mei betg clarifica cumpleinamein.}\\[0.3em]
\noindent{}Gemini 2.5 Flash:  \textit{Apparentamagn na funcziunescha betg gnieu iMessage sur WiFi, tgei ca na clerescha betg dal tut a mei.}\\
\subsubsection*{Surmiran}
\noindent{}\underline{Reference:} \textit{Scu para funcziunescha mianc iMessage sur WiFi, chegl tg'ia sa betg propi tgapeir}\\[0.3em]
\noindent{}Llama 3.3 (70B):  \textit{Apparentamain funcziuna betg ni ir iMessage sur WiFi, sche che m'èn betg del tut clar}\\[0.3em]
\noindent{}GPT-4o:  \textit{Apparentamaint funcziunescha gnanc iMessage sur WiFi, tge nun è dal tut cler per mai.}\\[0.3em]
\noindent{}Gemini 2.5 Flash:  \textit{Apparentamaintg na funcziunescha betg gniev iMessage sur WiFi, chegl tgi na m'è betg dal tuttafatg cler.}\\
\subsubsection*{Puter}
\noindent{}\underline{Reference:} \textit{Pera cha nu funcziuna niauncha iMessage sur WiFi, che ch'eau nun incleg dal tuot}\\[0.3em]
\noindent{}Llama 3.3 (70B):  \textit{Apparentamain funcziuna betg era neir iMessage sur WiFi, cha m'ha betg dal tut en persvasiun}\\[0.3em]
\noindent{}GPT-4o:  \textit{A para cha iMessage nu funcziuna nemain sur WiFi, quai nu'm para fich cler}\\[0.3em]
\noindent{}Gemini 2.5 Flash:  \textit{Apparentamaing nu funcziunescha neir betg iMessage sur WiFi, che nu'm clerescha dal tuot.}\\
\subsubsection*{Vallader}
\noindent{}\underline{Reference:} \textit{Apparaintamaing nu funcziuna gnanca üna jada iMessage sur WiFi, quai chi nu'm voul propcha ir per testa}\\[0.3em]
\noindent{}Llama 3.3 (70B):  \textit{Apparentamain funcziunia neir il iMessage sur WiFi, quai ch'ha mì nua betg dal tut en la testa}\\[0.3em]
\noindent{}GPT-4o:  \textit{Apparentamaing nu funcziuna neir iMessage sur WiFi, quai nu'm para fich cler.}\\[0.3em]
\noindent{}Gemini 2.5 Flash:  \textit{Apparentamaing nu funcziunescha neir betg iMessage sur WiFi, che nu'm para dal tuot cler.}\\
\subsection{Romansh to German}
\noindent{}[English:]  \textit{it seems like even iMessage over WiFi isn't working, which doesn't quite make sense to me}\\[0.3em]
\noindent{}\underline{German reference:} \textit{Anscheinend funktioniert nicht mal iMessage über WiFi, was mir nicht ganz einleuchtet}
\subsubsection*{Rumantsch Grischun}
\noindent{}Romansh source:  \textit{Para che gnanc iMessage funcziunia via WiFi, tge ch'è per mai betg dal tut evident.}\\[0.3em]
\noindent{}MADLAD-400 (10.7B) direct:  \textit{Damit iMessage über WLAN funktioniert, Ich weiß nicht, ob ich das jemals wiederholen werde.}\\[0.3em]
\noindent{}Supertext:  \textit{Anscheinend funktioniert iMessage nicht einmal über WiFi, was für mich nicht ganz selbstverständlich ist.}\\[0.3em]
\noindent{}Llama 3.3 (70B):  \textit{Dass nicht einmal iMessage über WiFi funktioniert, ist überhaupt nicht offensichtlich.}\\[0.3em]
\noindent{}GPT-4o:  \textit{Es scheint, dass nicht einmal iMessage über WiFi funktioniert, was für mich nicht ganz nachvollziehbar ist.}\\[0.3em]
\noindent{}Gemini 2.5 Flash:  \textit{Es scheint, dass nicht einmal iMessage über WLAN funktioniert, was für mich nicht ganz offensichtlich ist.}\\
\subsubsection*{Sursilvan}
\noindent{}Romansh source:  \textit{Sco ei para funcziunescha gnanc iMessage sur WiFi, quei ch'jeu sai buca propi capir.}\\[0.3em]
\noindent{}MADLAD-400 (10.7B) direct:  \textit{Er stellt fest, dass die Wii-Fernbedienung nicht funktioniert, und fragt sich, was er damit anfangen soll.}\\[0.3em]
\noindent{}Supertext:  \textit{Anscheinend funktioniert iMessage über WiFi nicht einmal, was ich wirklich nicht verstehen kann.}\\[0.3em]
\noindent{}Llama 3.3 (70B):  \textit{Weil ich als Funktion habe, dass ich iMessage über WiFi nutzen kann, was ich aber nicht richtig verstehe.}\\[0.3em]
\noindent{}GPT-4o:  \textit{Anscheinend funktioniert iMessage nicht einmal über WiFi, was ich wirklich nicht verstehen kann.}\\[0.3em]
\noindent{}Gemini 2.5 Flash:  \textit{Anscheinend funktioniert iMessage nicht einmal über WLAN, was ich nicht wirklich verstehen kann.}\\
\subsubsection*{Sutsilvan}
\noindent{}Romansh source:  \textit{Para funcziunescha gnànc iMessage sur igl WiFi, tge ca fa betga propi sen tanor me}\\[0.3em]
\noindent{}MADLAD-400 (10.7B) direct:  \textit{Die erste Singleauskopplung des Debütalbums, I Don’t Want to Miss a Thing, erreichte Platz 1 der US-Charts.}\\[0.3em]
\noindent{}Supertext:  \textit{Anscheinend funktioniert nicht einmal iMessage über WiFi, was auf jeden Fall nicht funktioniert.}\\[0.3em]
\noindent{}Llama 3.3 (70B):  \textit{Damit Funktionen wie iMessage über das WiFi funktionieren, was aber nicht gerade mein Ding ist}\\[0.3em]
\noindent{}GPT-4o:  \textit{Sogar iMessage funktioniert nicht über das WLAN, was für mich keinen Sinn ergibt.}\\[0.3em]
\noindent{}Gemini 2.5 Flash:  \textit{Anscheinend funktioniert iMessage nicht mal über WLAN, was für mich nicht wirklich Sinn macht}\\
\subsubsection*{Surmiran}
\noindent{}Romansh source:  \textit{Scu para funcziunescha mianc iMessage sur WiFi, chegl tg'ia sa betg propi tgapeir}\\[0.3em]
\noindent{}MADLAD-400 (10.7B) direct:  \textit{2000: 1. Preis beim Wettbewerb für Jugendtheaterstücke der Sächsischen Staatsoper Unter den Linden 2001: 1. Preis beim Wettbewerb für Jugendtheaterstücke der Sächsischen Staatsoper.}\\[0.3em]
\noindent{}Supertext:  \textit{Anscheinend funktioniert iMessage über WiFi nicht einmal, was man wirklich nicht verstehen kann.}\\[0.3em]
\noindent{}Llama 3.3 (70B):  \textit{Um Funktionen wie iMessage über WiFi zu nutzen, muss man wissen, dass das nicht ganz umsonst ist}\\[0.3em]
\noindent{}GPT-4o:  \textit{Anscheinend funktioniert iMessage nicht über WiFi, das kann ich nicht wirklich verstehen.}\\[0.3em]
\noindent{}Gemini 2.5 Flash:  \textit{Anscheinend funktioniert mein iMessage nicht über WLAN, was ich nicht wirklich verstehen kann.}\\
\subsubsection*{Puter}
\noindent{}Romansh source:  \textit{Pera cha nu funcziuna niauncha iMessage sur WiFi, che ch'eau nun incleg dal tuot}\\[0.3em]
\noindent{}MADLAD-400 (10.7B) direct:  \textit{Ich weiß nicht, ob iMessage über WLAN funktioniert, was ich nicht verstehe.}\\[0.3em]
\noindent{}Supertext:  \textit{Auch wenn iMessage über WiFi nicht funktioniert, ist das nicht ganz nachvollziehbar.}\\[0.3em]
\noindent{}Llama 3.3 (70B):  \textit{Denn nun funktioniert auch iMessage nicht mehr über WiFi, weil ich nicht mehr online bin}\\[0.3em]
\noindent{}GPT-4o:  \textit{Es scheint, dass iMessage nicht einmal über WiFi funktioniert, was ich überhaupt nicht verstehe.}\\[0.3em]
\noindent{}Gemini 2.5 Flash:  \textit{Aber dass iMessage auch über WLAN nicht funktioniert, das verstehe ich überhaupt nicht.}\\
\subsubsection*{Vallader}
\noindent{}Romansh source:  \textit{Apparaintamaing nu funcziuna gnanca üna jada iMessage sur WiFi, quai chi nu'm voul propcha ir per testa}\\[0.3em]
\noindent{}MADLAD-400 (10.7B) direct:  \textit{Anscheinend funktioniert iMessage nicht einmal mehr über WLAN, was mich sehr ärgert.}\\[0.3em]
\noindent{}Supertext:  \textit{Anscheinend funktioniert iMessage über WiFi nicht einmal, was mir wirklich nicht einleuchtet.}\\[0.3em]
\noindent{}Llama 3.3 (70B):  \textit{Aber offensichtlich funktioniert nicht einmal iMessage über WiFi, was mich nicht weiter wundert}\\[0.3em]
\noindent{}GPT-4o:  \textit{Anscheinend funktioniert iMessage nicht einmal über WiFi, was mir wirklich nicht in den Kopf will.}\\[0.3em]
\noindent{}Gemini 2.5 Flash:  \textit{Anscheinend funktioniert iMessage über WLAN nicht einmal, was mir wirklich nicht in den Kopf will.}\\

\end{document}